\documentclass[11pt,a4paper]{article}
\usepackage[hyperref]{acl2019}
\usepackage{times}
\usepackage{tabularx}
\usepackage{latexsym}
\usepackage{adjustbox}
\usepackage{multirow}
\usepackage{url}
\usepackage{xcolor}
\usepackage{hyperref}
\usepackage{lipsum}

\newcommand\blfootnote[1]{%
  \begingroup
  \renewcommand\thefootnote{}\footnote{#1}%
  \addtocounter{footnote}{-1}%
  \endgroup
}
\usepackage{colortbl}
\definecolor{mygray}{gray}{.9}
\aclfinalcopy 
\def\aclpaperid{***} 

\author{Xuewei Wang$^{*1}$, Weiyan Shi$^{*2}$, Richard Kim$^{2}$, Yoojung Oh$^2$\\ {\bf Sijia Yang$^{3}$, Jingwen Zhang$^2$ and Zhou Yu$^2$  }\\
~~\\$^1$ Zhejiang University,$^2$ University of California, Davis,$^3$ University of Pennsylvania
~~\\~~\\ cheriewang@zju.edu.cn,\{wyshi, khgkim, yjeoh\}@ucdavis.edu,\\
sijia.yang@asc.upenn.edu,\{jwzzhang, joyu\}@ucdavis.edu
}

\date{}

\begin{document}
\title{Persuasion for Good:\\
Towards a Personalized Persuasive Dialogue System for Social Good}

\maketitle
\begin{abstract}

   Developing intelligent persuasive conversational agents to change people's opinions and actions for social good is the frontier in advancing the ethical development of automated dialogue systems. To do so, the first step is to understand the intricate organization of strategic disclosures and appeals employed in human persuasion conversations. We designed an online persuasion task where one participant was asked to persuade the other to donate to a specific charity. We collected a large dataset with 1,017 dialogues and annotated emerging persuasion strategies from a subset. Based on the annotation, we built a baseline classifier with context information and sentence-level features to predict the 10 persuasion strategies used in the corpus. Furthermore, to develop an understanding of personalized persuasion processes, we analyzed the relationships between individuals' demographic and psychological backgrounds including personality, morality, value systems, and their willingness for donation. Then, we analyzed which types of persuasion strategies led to a greater amount of donation depending on the individuals' personal backgrounds. This work lays the ground for developing a personalized persuasive dialogue system.
   \blfootnote{* Equal contribution.} \footnote{The dataset and code are released at  \url{https://gitlab.com/ucdavisnlp/persuasionforgood}.}
   

\end{abstract}

\section{Introduction}

Persuasion aims to use conversational and messaging strategies to change one specific person's attitude or behavior. 
Moreover, personalized persuasion combines both strategies and user information related to the outcome of interest 
 to achieve better persuasion results \cite{1,2}.
 Simply put, the goal of personalized persuasion is to produce desired changes by making the information personally relevant and appealing. 
However, two questions about personalized persuasion 
still remain unexplored. First, we concern about how personal information would affect persuasion outcomes. Second, we question about what strategies are more effective considering different user backgrounds and personalities. 

The past few years have witnessed the rapid development of conversational agents. The primary goal of these agents is to facilitate task-completion and human-engagement in practical contexts \cite{7,8,9,10}. While persuasive technologies for behavior change have successfully leveraged other system features such as providing simulated experiences and behavior reminders \cite{11,12}, the development of automated persuasive agents remains lagged due to the lack of synergy between the social scientific research on persuasion and the computational development of conversational systems. 

In this work, we introduced the foundation work on building an automatic personalized persuasive dialogue system. We first collected 1,017  human-human persuasion conversations ({\scshape PersuasionForGood}) that involved real incentives to participants. Then we designed a persuasion strategy annotation scheme and annotated a subset of the collected conversations. In addition, we came to classify 10 different persuasion strategies using Recurrent-CNN with sentence-level features and dialogue context information. We also analyzed the  relations among participants' demographic backgrounds, personality traits, value systems, and their donation behaviors. Lastly, we analyzed what types of persuasion strategies worked more effectively for what types of personal backgrounds. These insights will serve as important elements during our design of the personalized persuasive dialogue systems in the next phase.

\section{Related Work}

In social psychology, the rationale for personalized persuasion comes from the Elaboration Likelihood Model (ELM) theory \cite{13}. It argues that people are more likely to engage with persuasive messages when they have the motivation and ability to process the information. The core assumption is that persuasive messages need to be associated with the  ways different individuals perceive and think about the world. Hence, personalized persuasion is not simply capitalizing on using superficial personal information such as name and title in the communication; rather, it requires a certain degree of understanding of the individual to craft unique messages that can enhance his or her motivation to process and comply with the persuasive requests \cite{1,2,3}. 

There has been an increasing interest in persuasion detection and prediction recently. \citet{hidey2017analyzing} presented a two-tiered annotation scheme to differentiate claims and premises, and different persuasion strategies in each of them in an online persuasive forum \cite{tan2016winning}. \citet{hidey2018persuasive} proposed to predict persuasiveness by modelling argument sequence in social media and showed promising results.  \citet{yang2019let} proposed a hierarchical neural network model to identify persuasion strategies in a semi-supervised fashion. Inspired by these prior work in online forums, we  present a persuasion dialogue dataset with user demographic and psychological attributes, and study personalized persuasion in a conversational setting. 

In the past few years, personalized dialogue systems have  come to people's attention because 
user-targeted personalized dialogue system is able to achieve better user engagement \cite{yu2016}. For instance, \citet{shi2018} exploited user sentiment information to make dialogue agent more user-adaptive and effective. But how to get access to user personal information is a limiting factor in personalized dialogue system design. \citet{persona2018} introduced a human-human chit-chat dataset with a set of 1K+ personas. In this dataset, each participant was randomly assigned a persona that consists of a few descriptive sentences. However, the brief description of user persona lacks quantitative analysis of users' sociodemographic backgrounds and psychological characteristics, and therefore is not sufficient for 
interaction effect analysis between personalities and  dialogue policy preference.

Recent research has advanced the dialogue system design on certain negotiation tasks such as bargain on goods \cite{hehe,e2e}. The difference between negotiation and persuasion lies in their ultimate goal. Negotiation strives to reach an agreement from both sides, while persuasion aims to change one specific person's attitude and decision. \citet{e2e} applied end-to-end neural models with self-play reinforcement learning to learn better negotiation strategies. In order to achieve different negotiation goals, \citet{hehe} decoupled the dialogue act and language generation which helped control the strategy with more flexibility. Our work is different in that we focus on the domain of persuasion and personalized persuasion procedure.

Traditional persuasive dialogue systems have been applied in different fields, such as law \cite{law}, car sales \cite{sale}, intelligent tutoring \cite{tutor}. However, most of them overlooked the power of  personalized design and didn't leverage deep learning techniques. 
Recently,  \citet{dataset1} considered personality traits in single-turn persuasion dialogues on social and political issues. They found that personality factors can affect belief change,  with conscientious, open and agreeable people being more convinced by emotional arguments. However, it's difficult to utilize such a single-turn dataset in the design of multi-turn dialogue systems. 
  

\begin{table*}[h]
\small
\begin{adjustbox}{width=\textwidth}
\begin{tabular}{ l| m{ 15cm} |l}
\hline
\bf Role & \bf Utterance & \bf Annotation \\
\rowcolor{mygray}
ER   & Hello, are you interested in protection of rights of children?       &  Source-related inquiry  \\
EE   & Yes, definitely. What do you have in mind?       &    \\\rowcolor{mygray}
ER   & There is an organisation called Save the Children and donations are essential to ensure children's rights to health, education and safety.       & Credibility appeal \\
EE   & Is this the same group where people used to "sponsor" a child?       &   \\\rowcolor{mygray}
ER   & Here is their website, https://www.savethechildren.org/. & Credibility appeal \\\rowcolor{mygray}
    & They help children all around the world. & Credibility appeal \\\rowcolor{mygray}
    & For instance, millions of Syrian children have grown up facing the daily threat of violence. & Emotion appeal \\\rowcolor{mygray}
    & In the first two months of 2018 alone, 1,000 children were reportedly killed or injured in intensifying violence.       &  Emotion appeal  \\
EE   & I can't imagine how terrible it must be for a child to grow up inside a war zone.        &   \\\rowcolor{mygray}
ER   & As you mentioned, this organisation has different programs, and one of them is to "sponsor" child. & Credibility appeal \\\rowcolor{mygray}
    & You choose the location.       & Credibility appeal \\
EE   & Are you connected with the NGO yourself?       &   \\\rowcolor{mygray}
ER   & No, but i want to donate some amount from this survey. & Self-modeling \\\rowcolor{mygray}
    & Research team will send money to this organisation.       &  Donation information  \\
EE   & That sounds great. Does it come from our reward/bonuses?       &   \\\rowcolor{mygray}
ER   & Yes, the amount you want to donate is deducted from your reward.       &   Donation information \\
EE   & What do you have in mind?       &   \\\rowcolor{mygray}
ER   & I know that my small donation is not enough, so i am asking you to also donate some small percentage from reward.       & Proposition of donation \\
EE   & I am willing to match your donation.       &   \\\rowcolor{mygray}
ER   & Well, if you go for full 0.30 i will have no moral right to donate less.       &   Self-modeling  \\
EE   & That is kind of you. My husband and I have a small NGO in Mindanao, Philippines, and it is amazing what a little bit of money can do to make things better.       & \\\rowcolor{mygray}
ER   & Agree, small amount of money  can mean a lot for people in third world countries. & Foot-in-the-door \\\rowcolor{mygray}
& So agreed? We donate full reward each??       &  Donation confirmation  \\
EE   & Yes, let's donate \$0.30 each. That's a whole lot of rice and flour. Or a whole lot of bandages.       &    \\\hline

\end{tabular}
\end{adjustbox}
\caption{\label{tb:0}An example  persuasion dialogue. ER and EE refer to the persuader and the persuadee respectively.}
\end{table*}

\section{Data Collection}

We designed an online persuasion task to collect emerging persuasion strategies from human-human conversations on the Amazon Mechanical Turk platform (AMT). We utilized ParlAI \cite{miller2017parlai}, a python-based platform that enables dialogue AI research, to assist the data collection. We picked \textit{Save the Children}\footnote{\url{https://www.savethechildren.org/}} as the charity to donate to, because it is one of the most well-known charity organizations around the world. 

Our task consisted of four parts, a pre-task survey, a persuasion dialogue, a donation confirmation and a post-task survey. Before the conversation began, we asked the participants to complete a pre-task survey to assess their psychological profile variables. There were four sub-questionnaires in our survey, the Big-Five personality traits \cite{Gold1992} (25 questions), the Moral Foundations endorsement \cite{Graham} (23 questions), the Schwartz Portrait Value (10 questions) \cite{Cie}, and the Decision-Making style (4 questions) \cite{Hami}. 
From the pre-task survey, we obtained a 23-dimension psychological feature vector where each element is the score of one characteristic, such as extrovert and agreeable.

Next, we randomly assigned the roles of persuader and persuadee to the two participants. The random assignment helped to eliminate the correlation between the persuader's persuasion strategies and the targeted persuadee's characteristics. In this task, the persuader needed to persuade the persuadee to donate part of his/her task earning to the charity, and the persuader could also choose to donate. Please refer to Fig.~\ref{fig:er_UI} and \ref{fig:ee_UI} in  Appendix for the data collection interface. For persuaders, we provided them with tips on different persuasion strategies along with some example sentences. 
For persuadees, they only knew they would talk about a specific charity in the conversation. Participants were encouraged to continue the conversation until an agreement was reached. Each participant was required to complete at least 10 conversational turns and multiple sentences in one turn were allowed. An example dialogue is shown in Table \ref{tb:0}. 

\begin{table}[h]
\small
\begin{adjustbox}{width=\columnwidth}
\begin{tabular}{llll}
\hline
\multicolumn{4}{l}{\textbf{Dataset Statistics}}                                        \\ \hline
\multicolumn{2}{l}{\# Dialogues}              & \multicolumn{2}{l}{1,017}              \\
\multicolumn{2}{l}{\# Annotated Dialogues ({\scshape AnnSet})}              & \multicolumn{2}{l}{300}              \\
\multicolumn{2}{l}{\#  Participants}          & \multicolumn{2}{l}{1,285}              \\
\multicolumn{2}{l}{Avg. donation}             & \multicolumn{2}{l}{\$0.35}             \\
\multicolumn{2}{l}{Avg. turns per dialogue}  & \multicolumn{2}{l}{10.43}              \\
\multicolumn{2}{l}{Avg. words per utterance}  & \multicolumn{2}{l}{19.36}              \\
\multicolumn{2}{l}{Total unique tokens}       & \multicolumn{2}{l}{8,141}            \\ \hline
\multicolumn{4}{l}{\textbf{Participants Statistics}}                                   \\ \hline
\textit{Metric}          & \textit{Persuader} & \multicolumn{2}{l}{\textit{Persuadee}} \\
Avg. words per utterance & 22.96              & \multicolumn{2}{l}{15.65}              \\
Donated           &   424 (42\%)        & \multicolumn{2}{l}{545 (54\%)}             \\
Not donated            &   593 (58\%)           & \multicolumn{2}{l}{472 (46\%)}             \\\hline

\end{tabular}
 \end{adjustbox}
\caption{\label{tb:sta1}Statistics of  {\scshape{PersuasionForGood}} }
\end{table}

After completing the conversation, both the persuader and the persuadee were asked to input the intended donation amount privately though a text box. The max amount of donation was the task payment. 
After the conversation ended, all participants were required to finish a post-survey assessing their sociodemographic backgrounds such as age and income. We also included several questions about their engagement in this conversation.

The data collection process lasted for two months and  the statistics of the collected dataset named {\scshape PersuasionForGood} are presented in Table \ref{tb:sta1}. We observed that on average persuaders chose to say longer utterances than persuadees (22.96 tokens compared to 15.65 tokens). During the data collection phase, 
we were glad to receive some positive comments from the workers. Some mentioned that it was one of the most meaningful tasks they had ever done on the AMT, which shows an acknowledgment to our task design.

\section{Annotation}

\begin{table}[h]
\begin{center}
\begin{tabular}{ l|l}
\hline \textbf{Category} & \textbf{Amount} \\ \hline
 Logical appeal  & 325 \\ 
Emotion appeal & 237 \\
Credibility appeal &  779 \\
Foot-in-the-door & 134 \\
Self-modeling & 150\\
Personal story & 91 \\
Donation information & 362  \\
Source-related inquiry & 167\\
Task-related inquiry & 180\\
Personal-related inquiry & 151\\
Non-strategy dialogue acts  & 1737  \\\hline
Total         & 4313 \\\hline
\end{tabular}
\end{center}
\caption{\label{tb:sum} Statistics of persuasion strategies in {\scshape AnnSet}.
}
\end{table}

After the data collection, we designed an annotation scheme to annotate different persuasion strategies  persuaders used. 
Content analysis method \cite{20} was employed  to create the annotation scheme. 
 Since our data was from typing conversation and the task was rather complicated, we observed that half of the conversation turns contained more than two sentences with different semantic meanings. So we chose to annotate each complete sentence instead of the whole conversation turn.

We also designed a dialogue act annotation scheme for persuadee's utterances, shown in Table~\ref{tab:persuadee act} in  Appendix, to capture persuadee's general conversation behaviors. We also recorded if the persuadee agreed to donate, and the intended donation amount mentioned in the conversation. 

We developed both persuader and persuadee's annotation schemes using theories of persuasion and a preliminary examination of 10 random conversation samples. Four research assistants independently coded 10 conversations, discussed disagreement, and revised the scheme accordingly. The four coders conducted two iterations of coding exercises on five additional conversations and reached an inter-coder reliability of Krippendorff's alpha of above 0.70 for all categories. Once the scheme was finalized, each coder separately coded the rest of the conversations. We named the 300 annotated conversations as the {\scshape AnnSet}. 

Annotations for persuaders' utterances included diverse argument strategies and task-related non-persuasive dialogue acts. Specifically, we identified 10 persuasion strategy categories that can be divided into two types, 1) \textbf{persuasive appeal} and 2) \textbf{persuasive inquiry}. Non-persuasive  dialogue acts included general ones such as greeting, and task-specific ones such as donation proposition and confirmation. Please refer to Table~\ref{tab:persuader act} in  Appendix for the persuader dialogue act scheme.


The seven strategies below belong to  \textbf{persuasive appeal}, which tries to change people's attitudes and decisions through different psychological mechanisms.

\noindent\textbf{Logical appeal} refers to the use of reasoning and evidence to convince others. For instance, a persuader can convince a persuadee that the donation will make a tangible positive impact for children using reasons and facts. 

\noindent\textbf{Emotion appeal} refers to the elicitation of specific emotions to influence others. Specifically, we identified four emotional appeals:~1) telling stories to involve participants, 2) eliciting empathy, 3) eliciting anger, and 4) eliciting the feeling of guilt.
\cite{guilty}.

\noindent\textbf{Credibility appeal} refers to the uses of credentials and citing organizational impacts to establish credibility and earn the persuadee's trust. The information usually comes from an objective source (\textit{e.g.}, the organization's website or other well-established websites).

\noindent\textbf{Foot-in-the-door}  refers to the strategy of starting with small donation requests to facilitate compliance followed by larger requests \cite{foot}. For instance, a persuader first asks for a smaller donation and extends the request to a larger amount after the persuadee shows intention to donate. 

\noindent\textbf{Self-modeling} refers to the strategy where the persuader first indicates his or her own intention to donate and chooses to act as a role model for the persuadee to follow. 

\noindent\textbf{Personal story} refers to the strategy of using narrative exemplars to illustrate someone's donation experiences or the beneficiaries' positive outcomes, which can motivate others to follow the actions.


\noindent\textbf{Donation information} refers to providing specific information about the donation task, such as the donation procedure, donation range, etc. By providing detailed action guidance, this strategy can enhance the persuadee's self-efficacy and facilitates behavior compliance.

The three strategies below belong to \textbf{persuasive inquiry}, which tries to facilitate more personalized persuasive appeals and to establish better interpersonal relationships by asking questions. 

\noindent\textbf{Source-related inquiry} asks if the persuadee is aware of the organization (i.e., the source in our specific donation task).

\noindent\textbf{Task-related inquiry} asks about the persuadee's opinion and expectation related to the task, such as their interests in knowing more about the organization.

\noindent\textbf{Personal-related inquiry} asks about the persuadee's previous personal experiences relevant to charity donation. 

The statistics of the {\scshape AnnSet} are shown in  Table \ref{tb:sum}, where we listed the number of times each persuasion strategy appears. Most of the further studies are  on the  {\scshape {AnnSet}}. Example sentences for each persuasion strategy are shown in Table~\ref{tab:sample sentence for strategy}.

\begin{table*}[h]
\small
\begin{adjustbox}{width=\textwidth}
\begin{tabular}{l|l}
\hline

\textbf{Persuasion Strategy}   & \textbf{Example} \\\hline
Logical appeal                 & \textit{\begin{tabular}[c]{@{}l@{}}Your donation could possible go to this problem and help many young children. \\ You should feel proud of the decision you have made today.\end{tabular}}                                     \\\hline
Emotion appeal               & \textit{\begin{tabular}[c]{@{}l@{}}Millions of children in Syria grow up facing the daily threat of violence.  \\ This should make you mad and want to help.\end{tabular}}            \\\hline

Credibility appeal             & \textit{\begin{tabular}[c]{@{}l@{}}And the charity is highly rated with many positive rewards.\\ You can find reports associated with the financial information by visiting this link.\end{tabular}}                             \\\hline

Foot-in-the-door      & \textit{\begin{tabular}[c]{@{}l@{}}And sometimes even a small help is a lot, thinking many others will do the same.\\ By people like you, making a a donation of just \$1 a day, you can feed a child for a month.\end{tabular}} \\\hline

Self-modeling     & \textit{\begin{tabular}[c]{@{}l@{}}I will  donate to Save the Children myself. \\ I will match your donation.\end{tabular}}                                                                                                      \\\hline
Personal story                 & \textit{\begin{tabular}[c]{@{}l@{}}I like to give a little money to charity each month. \\ My brother and I replaced birthday gifts with charity donations a few years ago.\end{tabular}}                                        \\\hline
Donation information & \textit{\begin{tabular}[c]{@{}l@{}}Your donation will be directly deducted from your task payment. \\ The research team will collect all donations and send it to Save the Children.\end{tabular}} \\\hline  

Source-related inquiry & \textit{\begin{tabular}[c]{@{}l@{}}Have you heard of Save the Children? \\ Are you familiar with the organization? \end{tabular}} \\\hline  

Task-related inquiry & \textit{\begin{tabular}[c]{@{}l@{}}Do you want to know the organization more? \\ What do you think of the charity?\end{tabular}} \\\hline  

Personal-related inquiry & \textit{\begin{tabular}[c]{@{}l@{}}Do you have kids? \\ Have you donated to  charity before?\end{tabular}} \\\hline  

\end{tabular}
\end{adjustbox}

\caption{ Example sentences for the 10 persuasion strategies.}
\label{tab:sample sentence for strategy}
\end{table*}

\begin{figure}[htb] 
 \center{\includegraphics[width=7cm]{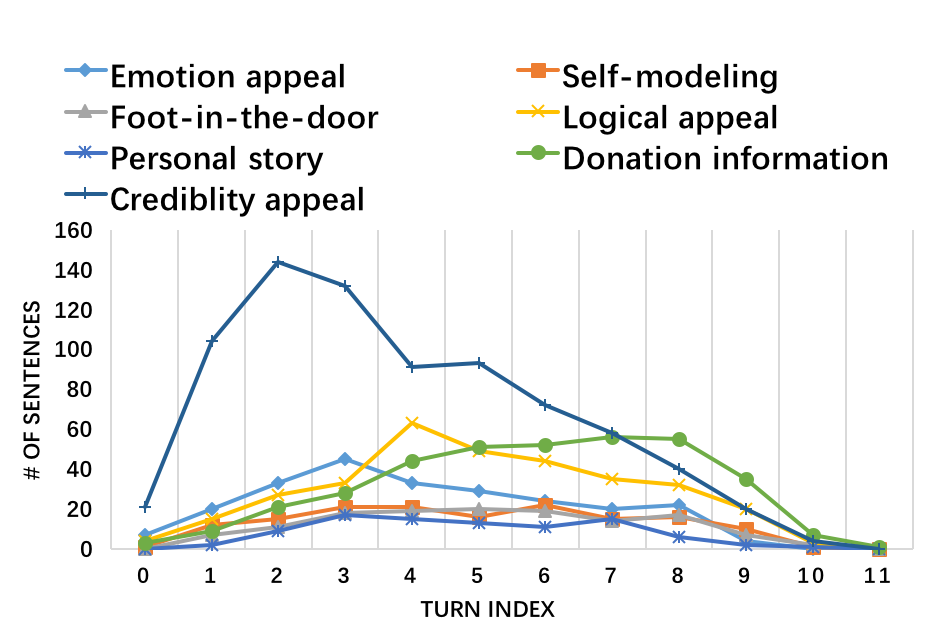}} 
 \caption{\label{fig:turn1} Distributions of the seven persuasive appeals across turns.} 
  \end{figure}

\begin{figure}[htb] 
  \center{\includegraphics[width=7cm]{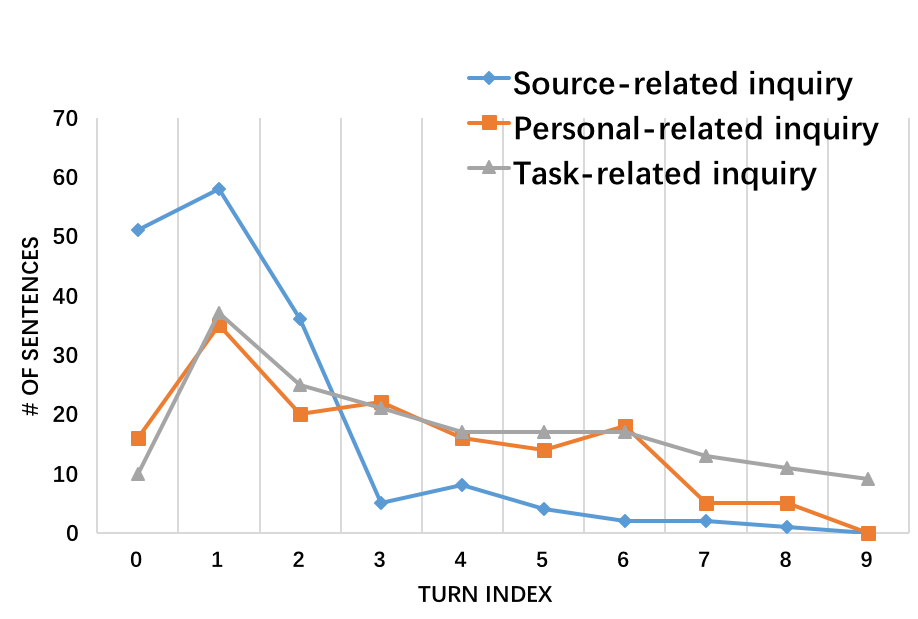}} 
  \caption{\label{fig:turn2} Distributions of the three persuasive inquiries across turns.} 
  \end{figure}

We first explored the distribution of different strategies across conversation turns. 
We present the number of different persuasion strategies at different conversation turn positions in Fig.~\ref{fig:turn1} (for persuasive appeal) and Fig.~\ref{fig:turn2} (for persuasive inquiry). 
As shown in Fig.~\ref{fig:turn1}, \textit{Credibility appeal} occurred more at the beginning of the conversations. In contrast, \textit{Donation information} occurred more in the latter part of the conversations. \textit{Logical appeal} and \textit{Emotion appeal} share a similar distribution and also frequently appeared in the middle of the conversations. The rest of the strategies,  \textit{Personal story}, \textit{Self-modeling} and \textit{Foot-in-the-door}, are spread out more evenly across the conversations, compared with the other strategies. For persuasive inquiries in Fig.~\ref{fig:turn2}, \textit{Source-related inquiry} mainly appeared in the first three turns, and the other two kinds of inquiries have a similar distribution.
 
 \section{Donation Strategy Classification}
 \begin{figure}[h]  
  \center{\includegraphics[width=0.85\columnwidth, height=5.4cm]  {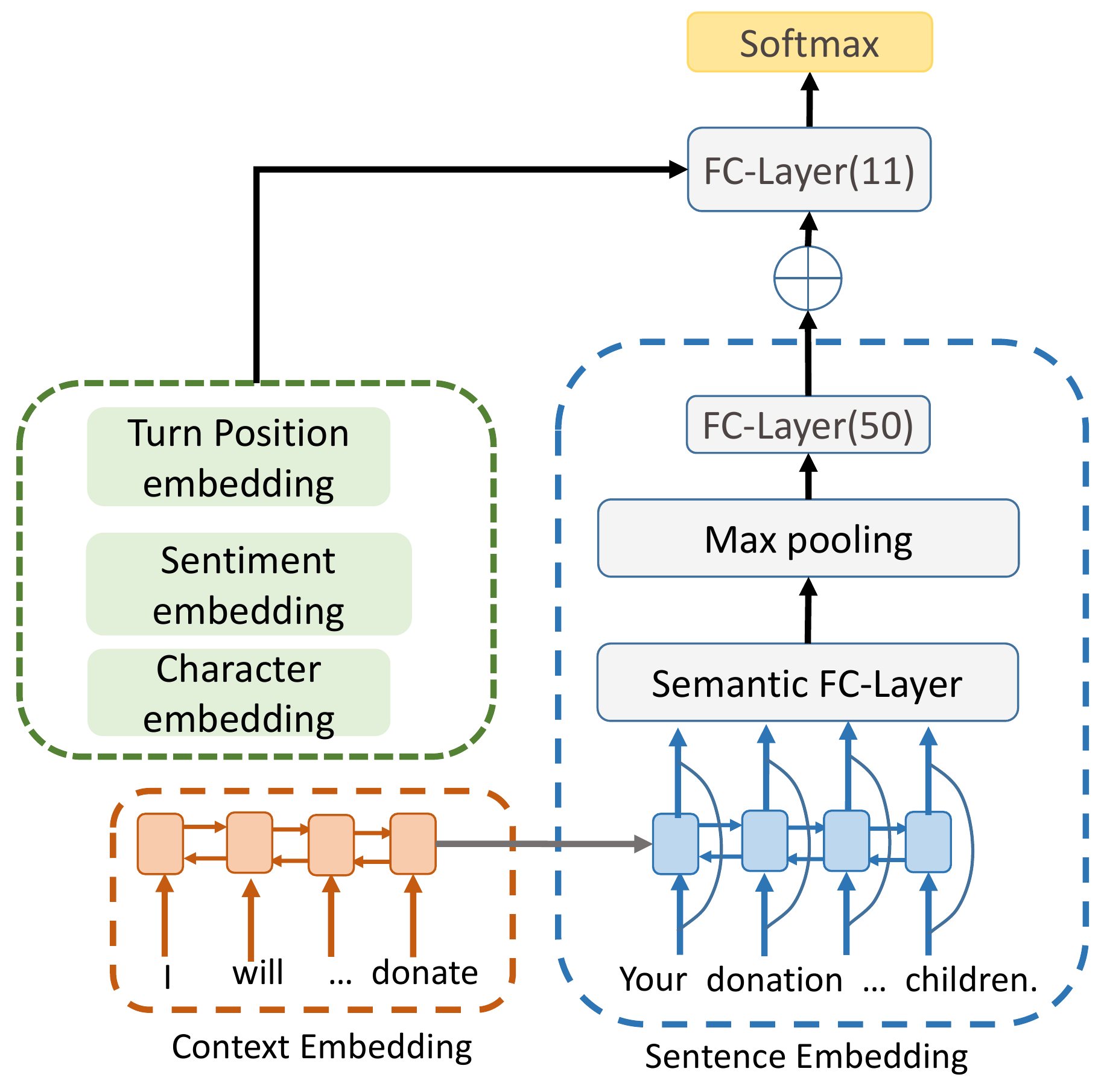}} 
  \caption{\label{fig:model} The hybrid RCNN model combines sentence embedding, context embedding and sentence-level features. ``+''  represents vector concatenation. 
  The blue dotted box shows the sentence embedding part. The orange dotted box shows the context embedding part. The green dotted box shows the sentence-level features.
  } 
  \end{figure}
 
In order to build a persuasive dialogue system, we need to first understand human persuasion patterns and differentiate various persuasion strategies. Therefore, 
we designed a classifier for the 10 persuasion strategies plus one additional ``non-strategy'' class for all the non-strategy dialogue acts in the {\scshape {AnnSet}}. 
We proposed a hybrid RCNN model which combined the following features, 1) sentence embedding, 2) context embedding and 3) sentence-level feature, for the classification. The model structure is shown in Fig.~\ref{fig:model}.

\noindent \textbf{Sentence embedding} used  recurrent convolutional neural network (RCNN), which combined CNN and RNN to extract both the global and local semantics, and the recurrent structure may reduce noise compared to the window-based neural network \cite{Lai2015}. We concatenated the word embedding and the hidden state of the LSTM as the sentence embedding $s_t$. Next, a linear semantic transformation was applied on $s_t$ to obtain the input to a max-pooling layer. Finally, the pooling layer was used to capture the effective information throughout the entire sentence. 

\noindent\textbf{Context embedding} was composed of the previous persuadee's utterance. Considering the relatively long context, we used the last hidden state of the context LSTM as the initial hidden state of the RCNN. We also experimented with other methods to extract context and will detail them in Section~6.

We also designed three \textbf{sentence-level features} 
to capture meta information other than embeddings. We describe them below. 

\noindent  \textbf{Turn position embedding.} According to the previous analysis, different strategies have different distributions across conversation turns, so the turn position may help the strategy classification. 
We condensed the turn position information into a 10-dimension embedding vector. 
  
\noindent \textbf{Sentiment.}
We also extracted sentiment features for each sentence using VADER \cite{vader}, a rule-based sentiment analyzer. It generates negative, positive, neutral scores from zero to one. It is interesting to note that for \textit{Emotion appeal}, the average negative sentiment score is 0.22, higher than the average positive sentiment score, 0.10. It seems negative sentiment words are used more frequently in \textit{Emotion appeal} because persuaders tend to describe sad facts to arouse empathy in \textit{Emotion appeal}. In contrast, positive words are used more frequently in \textit{Logical appeal}, 
because persuaders tend to describe more positive results from donation when using \textit{Logical appeal}.

\noindent  \textbf{Character embedding.} For short text, character level features can be helpful. \citet{charembed} utilized character embedding to improve the dialogue act classification accuracy. Following \citet{charembed}, we chose the pre-trained multiplicative LSTM (mLSTM) network on 80 million Amazon product reviews to extract 4096-dimension character-level features \cite{char}\footnote{\url{https://github.com/openai/generating-reviews-discovering-sentiment}}. Given the output character embedding, we applied a linear transformation layer with output size 50 to obtain the final character embedding.


\section{Experiments}


Because human-human typing conversations are complex, one sentence may belong to   multiple strategy categories; out of the concern for model simplicity, we chose to predict the most salient strategy for each sentence. Table~\ref{tb:sum} shows the dataset is highly imbalanced, so we used the macro F1 as the evaluation metric, in addition to  accuracy. 
We conducted five-fold cross validation, and used the average scores across folds to compare the performance of different models. We set the initial learning rate to be 0.001 and applied exponential decay every 100 steps. The training batch size was 32 and all models were trained for 20 epochs. In addition, dropout \cite{dropout} with a probability of 0.5 was applied to reduce over-fitting. We adopted the 300-dimension pre-trained FastText \cite{Fast} as word embedding. The RCNN model used a single-layer bidirectional LSTM with a hidden size of 200.
We describe two baseline models below for comparison.\\

\noindent\textbf{Self-attention BLSTM (BLSTM)} only considers a single-layer bidirectional LSTM with self-attention mechanism. After finetuning, we set the attention dimension to be 150.

\noindent\textbf{Convolutional neural network (CNN)}
 uses multiple convolution kernels to extract textual features. A softmax layer was applied in the end to generate the probability for each category. The hyperparameters in  the original implementation \cite{Kim2014} were used.


\subsection{Experimental Results}

\begin{table}[htb]
\small
\centering



    
\begin{tabular}{ lll}

\hline \textbf{Models} & \textbf{Accuracy}& \textbf{Macro F1} \\ \hline
Majority vote & 18.1\%  & 5.21\% \\
BLSTM + All features & 73.4\% & 57.1\% \\
CNN + All features & 73.5\% & 58.0\%\\

\hline
\multicolumn{3}{l}{\textbf{Hybrid RCNN with different features}} 
\\
\hline
Sentence only & 74.3\% & 59.0\% \\
Sentence + Context CNN & 72.5\%  & 54.5\%\\
Sentence + Context Mean & 74.0\% & 58.5\%\\
Sentence + Context RNN & 74.4\% & 59.3\%\\
Sentence + Context tf-idf & 73.5\% &57.6\%\\
    
Sentence + Turn position & 73.8\%& 59.4\%\\
Sentence + Sentiment & 73.6\% & 59.7\%\\ 
Sentence + Character & 74.5\% & 59.3\% \\
All features &  \textbf{74.8\%} & \textbf{59.6\%} \\\hline
\end{tabular}

\caption{\label{tb:com} All the features include sentence embedding, context embedding, turn position embedding, sentiment and character embedding. The hybrid RCNN model with all the features performed the best on the {\scshape {AnnSet}}. Baseline models in the upper section also used all the features but didn't perform as good as the hybrid RCNN. 
}
\end{table}

As shown in Table~\ref{tb:com}, the hybrid RCNN with all the features (sentence embedding, context embedding, turn position embedding, sentiment and character embedding) reached the highest accuracy (74.8\%) and F1 (59.6\%). 
Baseline models in the upper section of Table~\ref{tb:com} also used all the features but didn't perform as good as the hybrid RCNN.  We further performed ablation study on the hybrid RCNN to discover different features' impact on the model's performance. 
We experimented with four different context embedding methods, 1) CNN, 2) the mean of word embeddings, 3) RNN (the output of the RNN was the RCNN's initial hidden state), and 4) tf-idf. We found RNN achieved best result (74.4\%) and F1 (59.3\%). The experimental results  suggest incorporating context improved the model performance slightly but not significantly. This may be because in persuasion conversations,  sentences  are relatively long and contain complex semantic meanings, which makes it hard to encode the context information. This suggests we  develop better methods to extract important semantic meanings from the context in the future.
Besides, all three sentence-level features improved the model's F1. Although the sentiment feature only has three dimensions,  it still increased the model's F1 score. 

To further analyze the results, we plotted the confusion matrix for the best model in Fig.~\ref{fig:cm} in Appendix. We found the main error comes from the misclassification of \textit{Personal story}. Sometimes sentences of \textit{Personal story} were misclassified as \textit{Emotion appeal}, because a subjective story can contain sentimental words, which may confuse the model. 
Besides, \textit{Task-related inquiry}  was hard to classify due to the diversity of inquiries.
In addition, \textit{Foot-in-the-door} strategy can be mistaken for \textit{Logical appeal}, because when using \textit{Foot-in-the-door}, people would sometimes make logical arguments about  the small donation, such as describing the tangible effects of the small donation. 
For example, the sentence ``Even five cents can help save children's life.'' also mentioned the benefits from the small donation. Besides, certain sentences of \textit{Logical appeal} may contain emotional words, which led to the confusion between \textit{Logical appeal} and \textit{Emotion appeal}. In summary, due to the complex nature of  human-human typing dialogues, one sentence may convey multiple meanings, which led to  misclassifications.

\section{Donation Outcome Analysis}

After identifying and categorizing the persuasion strategies, the next step is to analyze the factors that contribute to the final donation decision. Specifically, understanding the effects of the persuader's strategies, the persuadee's personal backgrounds, and their interactions on donation can greatly enhance the conversational agent's capability to engage in personalized persuasion. 
Given the skewed distribution of intended donation amount from the persuadees, the outcome variable was dichotomized to indicate whether they donated or not (1 = making any amount of donation and 0 = none). Duplicate survey data from participants who did the task more than once were removed before the analysis, and for such duplicates, only data from the first completed task were retained. This pruning process resulted in an analytical sample of 252 unique persuadees in the {\scshape{ AnnSet}}. All measured demographic variables and psychological profile variables were entered into logistic models. Results are presented in Section \ref{sec:supplemental_analysis} 
in  Appendix. Our analysis consisted of three parts, including the effects of persuasion strategies on the donation outcome, the effects of persuadees' psychological backgrounds on the donation outcome, and the interaction effects among all strategies and personal backgrounds. 

\subsection{Persuasion Strategies and Donation}
Overall, among the 10 persuasion strategies,  \textbf{\textit{Donation information} showed a significant positive effect on the donation outcome} ($p < 0.05$), as shown in Table~\ref{tab:strategy} in  Appendix. This confirms previous research which showed efficacy information increases persuasion. More specifically, because \textit{Donation information} 
gives the persuadee  step-by-step instructions on how to donate, which makes the donation procedure more accessible and as a result, increases the donation probability. An alternative explanation is that persuadees with a strong donation intention were more likely to ask about the donation procedure, and therefore \textit{Donation information} appeared in most of the successful dialogues resulting in a donation.
These compounding factors led us to further analyze the effects of psychological backgrounds on the donation outcome. 

\subsection{Psychological Backgrounds and Donation}
We collected data on demographics and  four types of psychological characteristics, including moral foundation, decision style, Big-Five personality, and Schwartz Portrait Value, to analyze what types of people are more likely to donate and respond differently to different persuasive strategies. 

Results of the analysis on demographic characteristics in Table~\ref{tb:2} show that \textbf{the donation probability increases as the participant's age increases} ($p<0.05$). This may be due to the fact that older participants may have more money and may have children themselves, and therefore  are more willing to contribute to the children's charity. 
The Big-Five personality analysis shows that \textbf{more agreeable participants are more likely to donate ($p < 0.001$)};
the moral foundation analysis shows that \textbf{participants who care for others more have a higher probability for donation} ($p < 0.001$); the portrait value analysis shows that \textbf{participants who endorse benevolence more are also more likely to donate} ($p < 0.05$). 
These results suggest people who are more agreeable, caring about others, and endorsing benevolence are in general more likely to comply with the persuasive request  \cite{hoover2018moral, graham2013moral}.
On the decision style side, \textbf{participants who are rational decision makers are more likely to donate} ($p<0.05$), whereas intuitive decision makers are less likely to donate.

Another observation reveals participants' inconsistent donation behaviors. We found that some participants promised to donate during the conversation but reduced the donation amount or didn't donate at all in the end. In order to analyze these inconsistent behaviors, we selected the 236 persudees who agreed to donate in the {\scshape AnnSet}.  
Among these persuadees, 11\% (22) individuals reduced the actual donation amount and 43\% (88) individuals did not donate. Also, there are 3\% (7) individuals donated more than they mentioned in the conversation. We fitted the Big-Five traits score and the inconsistent behavior with a logistic regression model. The results in Table \ref{tb:inco} in  Appendix show that people who are more agreeable are more likely to match their words with their donation behaviors. But since the dataset is relatively small, the result is not significant and we should caution against overinterpreting these effects until we obtain more annotated data.  

 

 

\subsection{Interaction Effects of Persuasion Strategies and Psychological Backgrounds} 
To provide the necessary training data to build a personalized persuasion agent, we are interested in assessing not only the main effects of persuasion strategies employed by human persuaders, but more importantly, the presence of (or lack of) heterogeneity of such main effects on different individuals. In the case where the heterogeneous effects were absent, the  task of building the persuasive agent would be simplified because it wouldn't need to pay any attention to the targeted audience's attribute. Given the evidence shown in personalized persuasion, our expectation was to observe variations in the effects of persuasion strategies conditioned upon the persuadee's personal traits, especially the four psychological profile variables identified in the previous analysis (i.e., agreeableness, endorsement of care and benevolence, and rational decision making style).

Table~\ref{tab:inter-bigfive}, \ref{tab:inter-moral} and \ref{tab:inter-decision} present 
evidence for heterogeneity, conditioned upon the Big-Five personality traits, the moral foundation scores and the decision style. 
For example, although \textit{Emotion appeal} does not show a significant main effect averaged across all participants, it showed a significant positive effect on the donation probability of participants who are more extrovert ($p<0.05$). This suggests \textbf{when encountering  more extrovert persuadees, the agent can initiate \textit{Emotion appeal} more}. %

Besides, \textbf{\textit{Personal-related inquiry} significantly increases the donation probability of people who are more neurotic} ($p < 0.05$) in the Big-Five test, \textbf{but is negatively associated with the donation probability of people who endorse authority more}  in the moral foundation test. 
Given the relatively small dataset, we caution against overinterpreting these interaction effects until further confirmed after all the conversations in our dataset were content coded. With that said, the current set of evidence supports the presence of heterogeneity in the effects of persuasion strategies, which provide the basis for our next step to design a personalized persuasive system that aims to automatically identify and tailor persuasive messages to different individuals.

\section{Ethical Considerations}
Persuasion is a double-edged sword and has been used for good or evil throughout the history. Given the fast development of automated dialogue systems, an ethical design principle must be in place throughout all stages of the development and evaluation. As the Roman rhetorician Quintilian defined a persuader as ``a good man speaking well'', when developing persuasive agents, building an ethical and good intention that benefits the persuadees must come before designing and  engineering the conversational capability to persuade. For instance, we choose to use the donation task as a first step to develop a persuasive dialogue system because the relatively simple task involves persuasion to benefit children. Other persuasive contexts can consider designing persuasive agents to help individuals fulfill their goals such as engaging in more exercises or sustaining environmentally friendly actions. 
Second, when deploying the persuasive agents in real conversations, it is important to keep the persuadees informed of the nature of the dialogue system so they are not deceived. By revealing the identity of the persuasive agent, the persuadees need to have options to communicate directly with the human team behind the system. Similarly, the purpose of the collection of persuadees’ personal information and analysis on their psychological traits must be clearly communicated to the persuadees and the use of their data requires active consent procedure. 
Lastly, the design needs to ensure that the generated responses are appropriate and nondiscriminative. This requires continuous monitoring of the conversations to make sure the conversations comply with both universal and local ethical standards.

\section{Conclusions and Future Work}

A key challenge in persuasion study is the lack of high-quality data and the interdisciplinary research between computational linguistics and social science.
We proposed a novel persuasion task, and collected a rich human-human persuasion dialogue dataset with comprehensive user psychological study and persuasion strategy annotation. We have also shown that a classifier with three types of features (sentence embedding, context embedding and sentence-level features) can reach good results on persuasion strategy prediction. However, much future work is still needed to further improve the performance of the classifier, such as including more annotations and more dialogue context into the classification. Moreover, we found evidence about the interaction effects between psychological backgrounds and persuasion strategies. For example, when facing participants who are more open, we can consider using the \textit{Source-related inquiry} strategy. 
This project lays the groundwork for the next step, which is to design a user-adaptive persuasive dialogue system that can effectively choose appropriate strategies based on user profile information to increase the persuasiveness of the conversational agent.

\section*{Acknowledgments}
This work was supported by an Intel research gift. We thank Saurav Sahay, Eda Okur and Shachi Kumar for valuable discussions. We also thank many excellent Mechanical Turk contributors for building this dataset.

\bibliography{acl2019}
\bibliographystyle{acl_natbib}

\newpage
\appendix

\section{Appendices}
\label{sec:appendix}

\subsection{Annotation Scheme}
Table~\ref{tab:persuadee act} and \ref{tab:persuader act} show the annotation schemes for selected persuadee acts and persuader acts respectively. For the full annotation scheme, please refer to \url{https://gitlab.com/ucdavisnlp/persuasionforgood}.  
In the persuader's annotation scheme, there is a series of acts related to \textbf{persuasive proposition} (\textit{proposition of donation}, \textit{proposition of  amount}, \textit{proposition of confirmation}, and \textit{proposition of more donation}). In general, \textbf{proposition} is needed in persuasive requests because the persuader needs to clarify the suggested behavior changes. In our specific task, donation propositions have to happen in every conversation regardless of the donation outcome, and therefore is not influential on the final outcome. Further, its high frequency might dilute the results. Given these reasons, we didn't consider propositions as a strategy in our specific context. 

\begin{table}[h]
\centering

\begin{adjustbox}{width=0.9\columnwidth}
\begin{tabular}{ m{ 2.3cm}| m{ 4.2cm}}
\hline \textbf{Category} & \textbf{Description} \\ \hline
Ask org info & Ask questions about the charity   \\\hline
Ask donation procedure & Ask questions about how to donate   \\\hline
Positive reaction &  Express opinions/thoughts that may lead to a donation   \\\hline
Neutral reaction & Express opinions/thoughts neutral towards  a donation   \\\hline
Negative reaction & Express opinions/thoughts against a donation   \\\hline

Agree donation & Agree to donate \\\hline
Disagree donation  & Decline to donate\\\hline
Positive to inquiry & Show positive responses to persuader's inquiry \\\hline
Negative to inquiry & Show negative responses to persuader's inquiry \\\hline

\end{tabular}
\end{adjustbox}

\caption{\label{tab:persuadee act}  Descriptions of selected important \textbf{persuadee} dialogue acts. }
\end{table}

\begin{table}[h]
\centering

\begin{adjustbox}{width=0.9\columnwidth}
\begin{tabular}{ m{ 2.3cm}| m{ 4.2cm}}
\hline \textbf{Category} & \textbf{Description} \\ \hline
Proposition of donation & Propose donation   \\\hline
Proposition of  amount & Ask the specific donation amount   \\\hline
Proposition of confirmation &  Confirm donation   \\\hline
Proposition of more donation & Ask the persuadee to donate more   \\\hline
Experience affirmation & Comment on the persuadee's statements   \\\hline

Greeting  & Greet the persuadee\\\hline
Thank & Thank the persuadee \\\hline

\end{tabular}
\end{adjustbox}

\caption{\label{tab:persuader act}  Descriptions of selected important non-strategy \textbf{persuader} dialogue acts.}
\end{table}


%

%


\subsection{Donation Outcome Analysis Results}
\begin{figure}[htb!]
\centering
      \includegraphics[width=8cm]{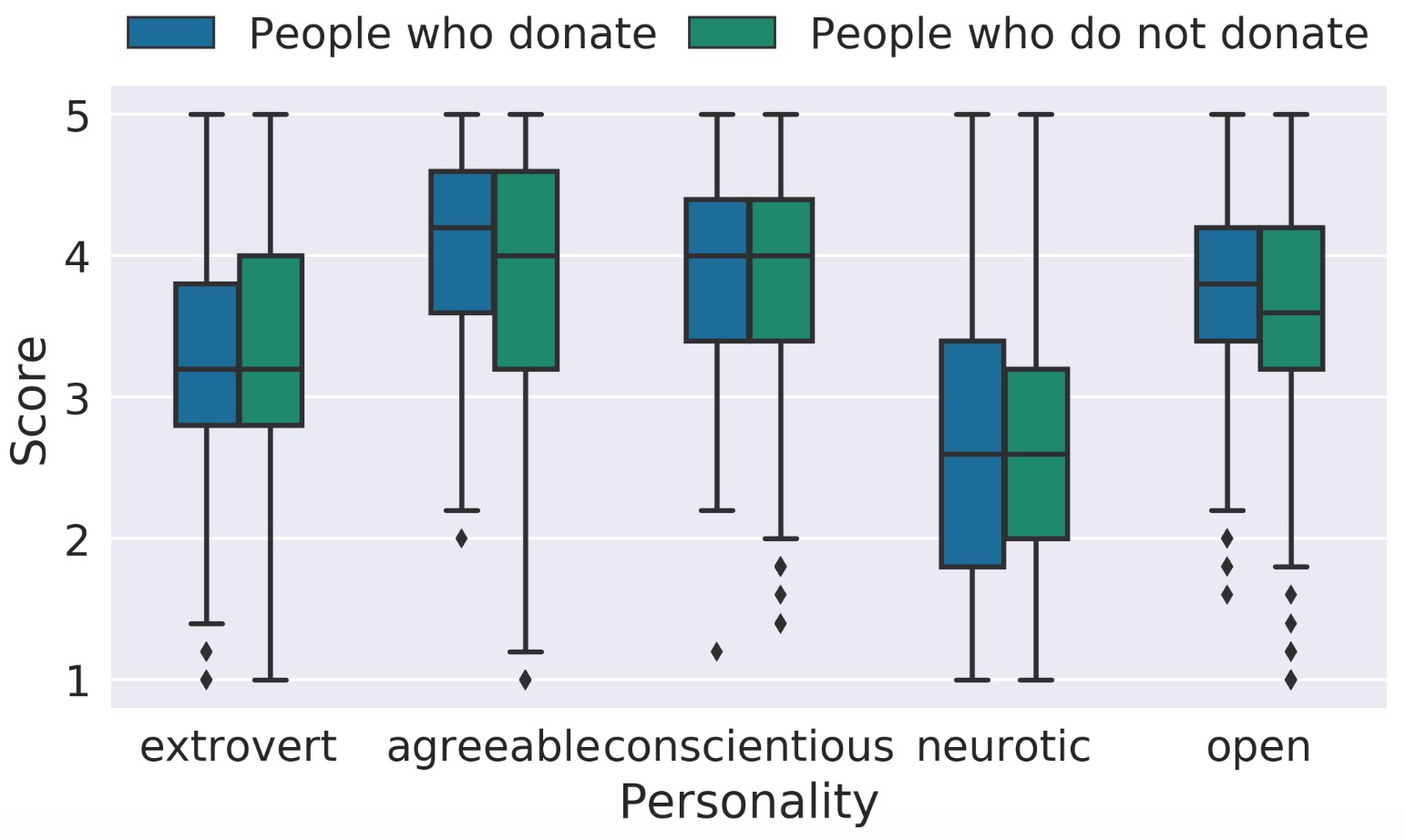}
        \caption{\label{fig:persona}\textbf{Big-Five traits score distribution for people who donated and didn't donate.} For all the 471 persuadees who did not donate in the {\scshape PersuasionForGood}, we compared their personalities score with the other 546 persuadees who donated. The result shows that people who donated have a higher score on agreeableness and openness in the Big-Five analysis. Because strategy annotation was not involved in the psychological analysis, we used the whole dataset (1017 dialogues) for this analysis.}
\label{fig:inconsistent and big-five}
 \end{figure}

We used {\scshape AnnSet} for the analysis except for Fig.~\ref{fig:inconsistent and big-five} and Table~\ref{tb:2}. Estimated coefficients of the logistic regression models predicting the donation probability (1 = donation, 0 =  no donation) with different variables are shown in Table~\ref{tab:strategy}, \ref{tb:inco},   \ref{tab:inter-decision}, \ref{tb:2}, \ref{tab:inter-bigfive}, and \ref{tab:inter-moral}. Two-tailed tests are applied for statistical significance where *$p < 0.05$, **$ p < 0.01$ and ***$ p < 0.001$  . 

\label{sec:supplemental_analysis}

\begin{table}[htb!]
\centering
\begin{tabular}{l|l}
\hline
\textbf{Persuasion Strategy}   & \textbf{Coefficient} \\ \hline
Logical appeal       & 0.06                 \\
Emotion appeal       & 0.03               \\
Credibility appeal   & -0.11                \\
Foot-in-the-door     & 0.06                 \\
Self-modeling        &  -0.02                    \\
Personal story       & 0.36                 \\
Donation information & 0.31*                 \\ 
Source-related inquiry       & 0.11                 \\
Task-related inquiry       & -0.004                 \\
Personal-related inquiry       & 0.02                 \\
\hline
\end{tabular}
\caption{\textbf{Associations between the persuasion strategies and the donation (dichotomized)}. *$p < 0.05$. {\scshape AnnSet} was used for the analysis.}
\label{tab:strategy}
\end{table}

\begin{table}[htb!]
\centering
\begin{tabular}{ l|l}
\hline
\textbf{Big-Five} &  \textbf{Coefficient}\\\hline
 extrovert  			& 0.22  \\
 agreeable  			&  -0.34 \\
 conscientious           & -0.27 \\
 neurotic                & -0.11   \\
 open                    & -0.19  \\\hline
\end{tabular}

\caption{
\label{tb:inco} \textbf{Associations between the Big-Five traits and the inconsistent donation behavior} (dichotomized, 1 = inconsistent donation behavior, 0 = consistent behavior). *$p < 0.05$. 
{\scshape AnnSet} was used for the analysis.}
\end{table}

\begin{table}[htb!]
\centering
\begin{tabular}{ll}
\hline
\textbf{Decision Style by \textit{Strategy}}                & \textbf{Coefficient}          \\ \hline
\textbf{Rational} by & \\
\textit{Logical appeal}         & -0.16                          \\
\textit{Emotion appeal}         & 0.35                          \\
\textit{Credibility appeal}     & -0.23                         \\
\textit{Foot-in-the-door}       & 0.41                      \\
\textit{Self-modeling}       & 0.19                      \\

\textit{Personal story}         & -0.32                          \\
\textit{Donation information}   & -0.32                         \\
\textit{Source-related inquiry}            & 0.36 \\
\textit{Task-related inquiry}            & 0.03 \\
\textit{Personal-related inquiry}            & 0.33 \\
\hline
\textbf{Intuitive} by & \\

\textit{Logical appeal}         & 0.01                          \\
\textit{Emotion appeal}         & 0.11                          \\
\textit{Credibility appeal}     & -0.04                         \\
\textit{Foot-in-the-door}       & 0.47*                      \\
\textit{Self-modeling}       & 0.13                      \\

\textit{Personal story}         & -0.31                          \\
\textit{Donation information}   & -0.02                         \\
\textit{Source-related inquiry}            & -0.29 \\
\textit{Task-related inquiry}            & 0.12 \\
\textit{Personal-related inquiry}            & 0.19 \\
\hline
\end{tabular}
\caption{\textbf{Interaction effects between decision style and the donation (dichotomized)}. *$p < 0.05$ . Coefficients of  the logistic regression predicting the donation probability (1 = donation, 0 =  no donation) are shown here. {\scshape AnnSet} was used for the analysis.}
\label{tab:inter-decision}
\end{table}

\begin{table}[htb!]
\centering
\begin{adjustbox}{width=7cm}
\begin{tabular}{ll}
\hline
\textbf{Predictor}                   & \textbf{Coefficient}\\ \hline
\multicolumn{2}{l}{\textbf{Demographics}}                                                                                      \\
Age                                  & 0.02*\\
Sex: Male vs. Female                 & -0.11\\
Sex: Other vs. Female                & -0.14\\
Race: White vs. Other                & 0.28\\
Less Than Four-Year College vs.      & \multirow{2}{*}{0.16}\\
Four-Year College                    &\\
Postgraduate vs. Four-Year College   & -0.20\\
Marital: Unmarried vs. Married       & -0.21\\
Employment: Other vs. Employed       & 0.17\\
Income (continuous)                  & -0.01\\
Religion: Catholic vs. Atheist       & 0.34\\
Religion: Other Religion vs. Atheist & 0.21\\
Religion: Protestant vs. Atheist     & 0.15\\
Ideology: Liberal vs. Conservative   & 0.11\\
Ideology: Moderate vs. Conservative  & -0.04\\ \hline
\multicolumn{2}{l}{\textbf{Big-Five Personality Traits}}                                                                       \\
Extrovert                            & -0.17\\
Agreeable                            & 0.58***\\
Conscientious                        & -0.15\\
Neurotic                             & 0.09                 \\
Open                                 & -0.01\\ \hline
\multicolumn{2}{l}{\textbf{Moral Foundation}}                                                            \\
Care/Harm                            & 0.38***\\
Fairness/Cheating                    & 0.08\\
Loyalty/Betrayal                     & 0.09\\
Authority/Subversion                     & 0.04\\
Purity/Degradation                   & -0.02\\
Freedom/Suppression                  & -0.13\\ \hline
\multicolumn{2}{l}{\textbf{Schwartz Portrait Value}}                                                     \\
Conform                              & -0.07\\
Tradition                            & 0.06\\
Benevolence                          & 0.18*\\
Universalism                         & 0.05\\
Self-Direction                       & -0.06\\
Stimulation                          & -0.08\\
Hedonism                             & -0.10\\
Achievement                          & -0.03\\
Power                                & -0.05\\
Security                             & 0.09\\ \hline
\multicolumn{2}{l}{\textbf{Decision-Making Style}}                                                                             \\
Rational                             & 0.25*\\
Intuitive                            & -0.02\\
\hline
\end{tabular}

\end{adjustbox}
\caption{
\label{tb:2} \textbf{Associations between the psychological profile and the donation (dichotomized)}. *$p < 0.05$, ***$p < 0.001$ . Estimated coefficients from a logistic regression predicting the donation probability ((1 = donation, 0 =  no donation)) are shown here. Because strategy annotation is not involved in the demographical and psychological analysis, we used the whole dataset (1017 dialogues) for this analysis.} 
\end{table}

\begin{table}[h!]
\centering
\small
\begin{tabular}{ll}
\hline

\textbf{Big-Five by \textit{Strategy}}                & \textbf{Coefficient}          \\ \hline
\textbf{Extrovert} by & \\
\textit{Logical appeal}         & -0.13                          \\
\textit{Emotion appeal}         & 0.54*                          \\
\textit{Credibility appeal}     & 0.08                         \\
\textit{Foot-in-the-door}       & 0.05                      \\
\textit{Self-modeling}       & -0.25                      \\

\textit{Personal story}         & -0.37                          \\
\textit{Donation information}   & -0.20                         \\
\textit{Source-related inquiry}            & -0.03 \\
\textit{Task-related inquiry}            & -0.49 \\
\textit{Personal-related inquiry}            & 0.43 \\

\hline
\textbf{Agreeable} by & \\

\textit{Logical appeal}         & -0.05                          \\
\textit{Emotion appeal}         & 0.34                          \\
\textit{Credibility appeal}     & 0.19                         \\
\textit{Foot-in-the-door}       & -0.04                      \\
\textit{Self-modeling}       & -0.68                      \\

\textit{Personal story}         & 0.50                          \\
\textit{Donation information}   & -0.10                         \\
\textit{Source-related inquiry}            & -1.34* \\
\textit{Task-related inquiry}            & -0.82* \\
\textit{Personal-related inquiry}            & 0.06 \\
\hline
\textbf{Neurotic} by & \\

\textit{Logical appeal}         & 0.43*                          \\
\textit{Emotion appeal}         & 0.30                          \\
\textit{Credibility appeal}     & -0.20                         \\
\textit{Foot-in-the-door}       & 0.38                      \\
\textit{Self-modeling}       & -0.38                      \\

\textit{Personal story}         & -0.70                          \\
\textit{Donation information}   & 0.22                         \\
\textit{Source-related inquiry}            & -0.29 \\
\textit{Task-related inquiry}            & -0.01 \\
\textit{Personal-related inquiry}            & 0.76* \\
\hline
\textbf{Open}  by& \\

\textit{Logical appeal}         & 0.48                          \\
\textit{Emotion appeal}         & 0.71                         \\
\textit{Credibility appeal}     & -0.13                         \\
\textit{Foot-in-the-door}       & -1.14                      \\
\textit{Self-modeling}       & 0.37                      \\

\textit{Personal story}         & -0.05                          \\
\textit{Donation information}   & -0.15                         \\
\textit{Source-related inquiry}            & 1.40 \\
\textit{Task-related inquiry}            & 0.70 \\
\textit{Personal-related inquiry}            & 0.24 \\
\hline
\textbf{Conscientious} by &  \\

\textit{Logical appeal}         & 0.16                         \\
\textit{Emotion appeal}         & 0.36                          \\
\textit{Credibility appeal}     & -0.58*                         \\
\textit{Foot-in-the-door}       & 1.22                      \\
\textit{Self-modeling}       & -0.12                      \\

\textit{Personal story}         & -1.47                          \\
\textit{Donation information}   & 0.70                         \\
\textit{Source-related inquiry}            & 0.23 \\
\textit{Task-related inquiry}            & -0.002 \\
\textit{Personal-related inquiry}            & 0.47 \\
\hline

\end{tabular}
\caption{\textbf{Interaction effects between Big-Five personality scores and the donation (dichotomized)}. *$p < 0.05$, **$ p < 0.01$. Coefficients of the logistic regression predicting the donation probability (1 = donation, 0 =  no donation) are shown here. {\scshape AnnSet} was used for the analysis.\newline \\~\\}
\label{tab:inter-bigfive}
\end{table}

\begin{table}[h!]
\centering
\small
\begin{tabular}{ll}
\hline

\textbf{Moral Foundation by \textit{Strategy}}                & \textbf{Coefficient}          \\ \hline
\textbf{Care} by & \\
\textit{Logical appeal}         & -0.03                          \\
\textit{Emotion appeal}         & -0.07                          \\
\textit{Credibility appeal}     & 0.26                         \\
\textit{Foot-in-the-door}       & -0.33                      \\
\textit{Self-modeling}       & 0.26                      \\

\textit{Personal story}         & 0.08                          \\
\textit{Donation information}   & -0.47                         \\
\textit{Source-related inquiry}            & 0.17 \\
\textit{Task-related inquiry}            & -0.38 \\
\textit{Personal-related inquiry}            & 0.96 \\

\hline
\textbf{Fairness} by & \\

\textit{Logical appeal}         & 0.35                          \\
\textit{Emotion appeal}         & 0.07                          \\
\textit{Credibility appeal}     & 0.08                         \\
\textit{Foot-in-the-door}       & 0.60                      \\
\textit{Self-modeling}       & 0.37                      \\

\textit{Personal story}         & -0.84                          \\
\textit{Donation information}   & 0.13                         \\
\textit{Source-related inquiry}            & 1.19 \\
\textit{Task-related inquiry}            & 0.52 \\
\textit{Personal-related inquiry}            & -0.69 \\
\hline
\textbf{Loyalty} by & \\

\textit{Logical appeal}         & -0.07                          \\
\textit{Emotion appeal}         & -0.07                          \\
\textit{Credibility appeal}     & 0.23                         \\
\textit{Foot-in-the-door}       & 0.40                      \\
\textit{Self-modeling}       & -0.01                      \\

\textit{Personal story}         & -0.23                          \\
\textit{Donation information}   & -0.31                         \\
\textit{Source-related inquiry}            & 0.70 \\
\textit{Task-related inquiry}            & -0.14 \\
\textit{Personal-related inquiry}            & -0.02 \\
\hline
\textbf{Authority}  by& \\

\textit{Logical appeal}         & 0.35                          \\
\textit{Emotion appeal}         & -0.15                         \\
\textit{Credibility appeal}     & -0.03                         \\
\textit{Foot-in-the-door}       & -0.83                      \\
\textit{Self-modeling}       & 0.39                      \\

\textit{Personal story}         & -0.41                          \\
\textit{Donation information}   & -0.27                         \\
\textit{Source-related inquiry}            & 0.11 \\
\textit{Task-related inquiry}            & -0.52 \\
\textit{Personal-related inquiry}            & -0.97* \\
\hline
\textbf{Purity} by &  \\

\textit{Logical appeal}         & -0.33                         \\
\textit{Emotion appeal}         & 0.22                          \\
\textit{Credibility appeal}     & -0.30*                         \\
\textit{Foot-in-the-door}       & 0.19                      \\
\textit{Self-modeling}       & -0.40                      \\

\textit{Personal story}         & 0.33                          \\
\textit{Donation information}   & 0.39                         \\
\textit{Source-related inquiry}            & -1.00* \\
\textit{Task-related inquiry}            & 0.29 \\
\textit{Personal-related inquiry}            & 0.29 \\
\hline
\textbf{Freedom} by& \\

\textit{Logical appeal}         & -0.02                          \\
\textit{Emotion appeal}         & 0.33                          \\
\textit{Credibility appeal}     & -0.33*                        \\
\textit{Foot-in-the-door}       & -0.37                      \\
\textit{Self-modeling}       & -0.09                      \\

\textit{Personal story}         & 0.06                          \\
\textit{Donation information}   & -0.02                         \\
\textit{Source-related inquiry}            & -0.41 \\
\textit{Task-related inquiry}            & -0.22 \\
\textit{Personal-related inquiry}            & 0.68 \\
\hline

\end{tabular}
\caption{\textbf{Interaction effects between moral foundation and the donation (dichotomized)}. *$p < 0.05$. 
}
\label{tab:inter-moral}
\end{table}



\subsection{Classification Confusion Matrix}
\begin{figure*}[h] 
 \center{\includegraphics[width=12cm]  {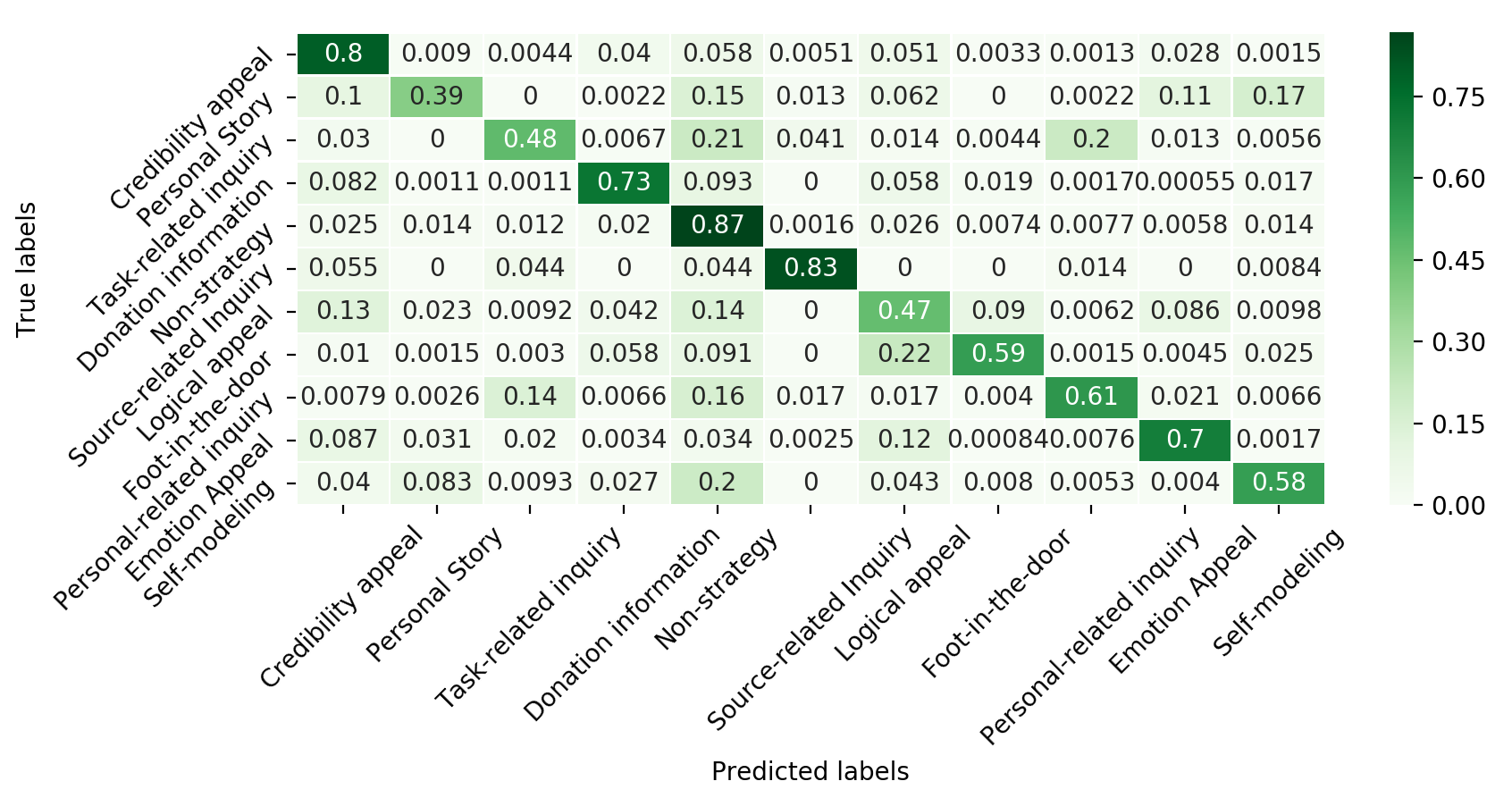}} 
 \caption{\label{fig:cm} Confusion matrix for the ten persuasion strategies and the non-strategy category  on the {\scshape AnnSet} using the hybrid RCNN model with all the  features.} 
 \label{fig:confusion matrix}
 \end{figure*}

Fig.~\ref{fig:confusion matrix} shows the classification confusion matrix.

\subsection{Data Collection Interface}
\begin{figure*}[htb!] 
 \center{\includegraphics[width=14cm]  {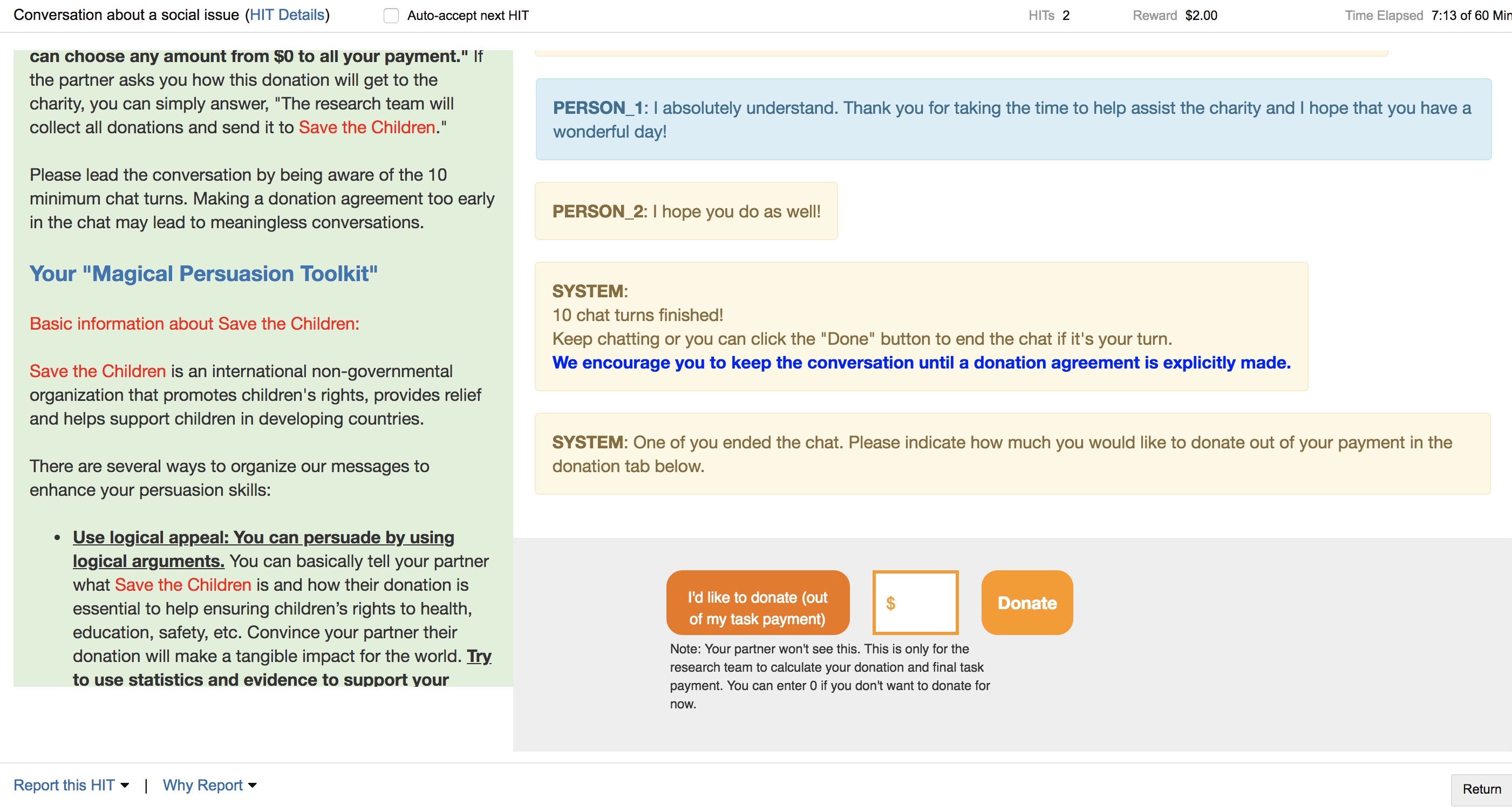}} 
 \caption{\label{fig:er_UI} Screenshot of the persuader's chat interface} 
 \end{figure*}

\begin{figure*}[htb!] 
 \center{\includegraphics[width=14cm]  {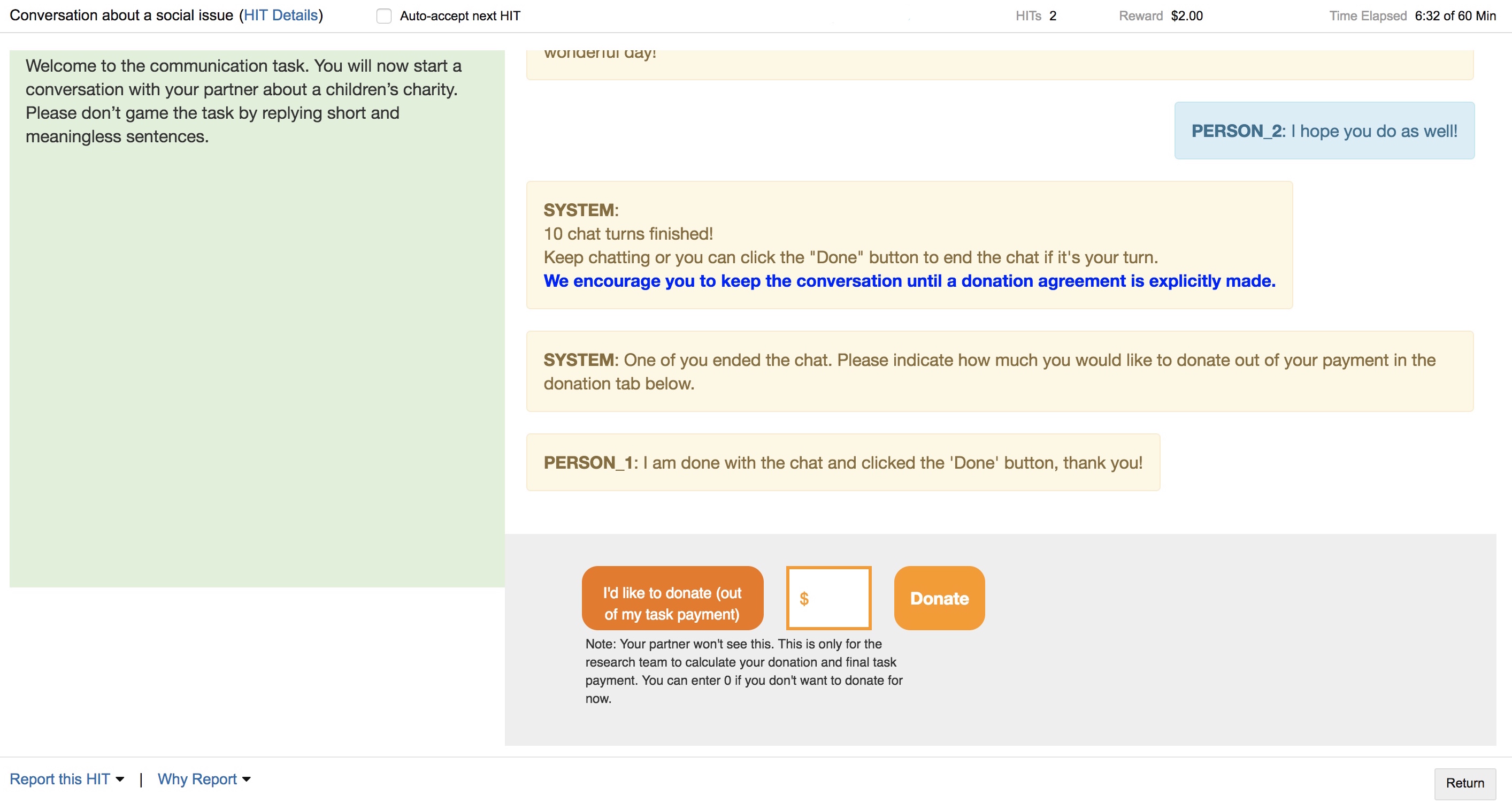}} 
 \caption{\label{fig:ee_UI} Screenshot of the persuadee's chat interface} 
 \end{figure*}

Fig.~\ref{fig:er_UI} and \ref{fig:ee_UI} shows the data collection interface.


\end{document}